# BUILDING HUMAN-LIKE COMMUNICATIVE INTELLIGENCE:
# A GROUNDED PERSPECTIVE.

## MARINA DUBOVA

Cognitive Science Program, Indiana University, 1101 E. 10th Street, Bloomington, IN 47405 USA

**Author Note**

Marina Dubova  https://orcid.org/0000-0001-5264-0489

I have no known conflict of interest to disclose.

Correspondence concerning this article should be addressed to Marina Dubova,

Cognitive Science Program, Indiana University, 1101 E. 10th Street, Bloomington, IN 47405 USA

Contact: mdubova@iu.edu

# ABSTRACT

Modern Artificial Intelligence (AI) systems excel at diverse tasks, from image classification to strategy games, even outperforming humans in many of these domains. After making astounding progress in language learning in the recent decade, AI systems, however, seem to approach the ceiling that does not reflect important aspects of human communicative capacities. Unlike human learners, communicative AI systems often fail to systematically generalize to new data, suffer from sample inefficiency, fail to capture common-sense semantic knowledge, and do not translate to real-world communicative situations. Cognitive Science offers several insights on how AI could move forward from this point. This paper aims to: (1) suggest that the dominant cognitively-inspired AI directions, based on nativist and symbolic paradigms, lack necessary substantiation and concreteness to guide progress in modern AI, and (2) articulate an alternative, "grounded", perspective on AI advancement, inspired by Embodied, Embedded, Extended, and Enactive Cognition (4E) research. I review results on 4E research lines in Cognitive Science to distinguish the main aspects of naturalistic learning conditions that play causal roles for human language development. I then use this analysis to propose a list of concrete, implementable components for building "grounded" linguistic intelligence. These components include embodying machines in a perception-action cycle, equipping agents with active exploration mechanisms so they can build their own curriculum, allowing agents to gradually develop motor abilities to promote piecemeal language development, and endowing the agents with adaptive feedback from their physical and social environment. I hope that these ideas can direct AI research towards building machines that develop human-like language abilities through their experiences with the world.

*Keywords*: artificial intelligence, language acquisition, communication, cognitively-inspired AI, embodied cognition

# 1. INTRODUCTION

The ability to acquire and use rich communicative systems (natural languages) is one of the crucial achievements of human minds. Natural language acquisition is, arguably, an intellectual capability that distinguishes us not only from other species, but also from machines. In this paper, I argue that rich, controllable, adaptive, and multi-agent environmental settings are crucial for language acquisition in humans. I use these insights to articulate a set of concrete cognitively-inspired directions for building machines that can develop human-like language abilities.

For decades, language has been one of the most challenging domains for AI to tackle. The recent wave of end-to-end neural network systems led to an impressive progress in language translation, analysis, and understanding. Despite this massive achievement, modern AI systems have been argued to lack fundamental properties that are necessary to achieve human-level performance in linguistic communication.

First, current state-of-the-art language models are **not adapted to the open-ended settings**, which is a fundamental property of human communicative interactions. Marcus and Davis (2019) argue that humans deal with completely novel situations on a day-to-day basis, whereas AI models are rarely prepared for unboundedly new patterns of data. For example, language models often fail to generalize to the examples that are outside of the distributions that they were trained on (e.g. Mccoy, Min, & Linzen, 2020) and these models are typically not equipped with the mechanisms to update their knowledge after the training period has passed (as described in Moskvichev & Liu, 2021). These limitations challenge the models' ability to succeed in the tasks that involve interacting with humans in the naturalistic situations which require constant adaptations of the communicating partners to each other and to the continuously changing environment.

Second, most artificial language systems are trained on **extremely large supervising datasets** (with the correct answers provided for examples), which are typically abstracted away from any perceptual modality. Despite receiving massive amounts of supervising training data, state-of-the-art artificial language systems often fail to generate good predictions for word and sentence patterns that have not been well represented in the training data (e.g. Lake & Baroni, 2018; Barrett et al., 2018; Glockner, Shwartz, & Goldberg, 2018; McCoy, Min, & Linzen, 2020; Gardner et al., 2020). In contrast, humans infer the meanings of new words and sentences from the context and previous knowledge, and efficiently generate new composite meanings out of the already learned ones in almost every communicative interaction.

Perhaps as a consequence of amodal learning, modern artificial language systems are often **incapable of capturing "common-sense"** semantic knowledge, which includes numbers and magnitudes, possible states of the world, causal relationships between entities in the environment, and many other components (Marcus & Davis, 2019; Lake et al., 2017), which often drive linguistic inferences in humans. For example, the state-of-the-art AI language models GPT-2 (Radford et al., 2019) and GPT-3 (Brown et al., 2020), trained on 40GB and 45TB textual data from the Internet respectively, produce syntactically sound sentences, but the contents of these sentences sometimes make no sense to human readers. Consider these examples of continuations suggested by GPT-2 (1) and GPT-3 (2) (in bold) noted by Marcus and Davis (Marcus, 2020a; Marcus & Davis, 2020):

*1) I have four plates and put one cookie on each. The total number of cookies is **24, 5 as a topping and 2 as the filling***

*2) You are having a small dinner party. You want to serve dinner in the living room. The dining room table is wider than the doorway, so to get it into the living room, you will have to **remove the door. You have a table saw, so you cut the door in half and remove the top half.***

For decades, the dominant strategy in the AI community has been to use bigger models and bigger data to cure these failures. For example, whereas GTP-2 had 1.5 billion adjustable parameters and was trained on 10 billion tokens, the newer GPT-3 model has 175 billion parameters and was trained on 499 billion tokens. While increasing the size of the model and the size of training set led to improvements, it seems evident that this approach might not solve the fundamental challenges of the language acquisition problem, as described below. As we will see, taking a "grounded" cognition perspective on language learning requires taking a different approach. Rather than focusing on bigger models and bigger data, I argue that researchers should revise the fundamental properties of these datasets, models, and their training regimes to allow agents to gradually develop language in rich, embodied, and adaptive conditions.

The increasing dissatisfaction with AI progress in the domain of language acquisition has led engineers and theoreticians to look for guiding insights from theories and empirical studies on the mechanisms of human learning. Below, I sketch out two most popular directions of integrating cognitive science and AI (nativist and neuro-symbolic), and argue that these approaches will ultimately fail to produce machines with human-like language abilities. I then argue for a grounded (4E) perspective on cognitively-inspired language learning in AI.

**Who this paper is for:** this paper aims to provide guiding insights to AI researchers who strive to develop systems with human-like language abilities. Even though human-like language abilities are desirable for some applications of communicative AI, they are not needed in most of them. Current state-of-the-art communicative AI models (e.g. GPT-2 and GPT-3) could serve very well for most applications, and I am fascinated by how far language learning through amodal and passive experiences has gone. This paper sketches out potential directions that scientists and engineers could explore in order to produce the communicative systems that share human conceptual knowledge, can participate in, and efficiently learn from the real-world communicative interactions.

## 2. Cognitively-Inspired AI: Directions

In this section, I briefly review and critically evaluate the two proposed solutions for building machines with human-like language abilities, which are largely inspired by nativist and symbolic perspectives in Cognitive Science. Then, I introduce an alternative, *grounded*, view on building human-like AI.

### 2.1. Core Knowledge, Language Acquisition Device, and the Building Blocks of Human Intelligence

Cognitive scientists have long been interested in discovering the core cognitive components that underlie human intelligence. Some researchers have attempted to discover "innate" (unlearned) cognitive capabilities, whereas others have tried to isolate the components that are the most essential for cognitive development while remaining agnostic about their origins. The core knowledge framework has been one source of inspiration as researchers attempt to build intelligent machines. According to this view, human and animal intelligence builds on a collection of specialized, domain-specific systems for representing objects, actions, number, space, and social agents (e.g. Spelke & Kinzler, 2006). Following these ideas, it has been suggested that computer scientists should focus on either hardwiring innate components into artificial brains or emphasizing the hypothesized "core" building blocks of intelligence during model training (Lake et al., 2017; Marcus & Davis, 2019; Marcus, 2020b; Chollet, 2019). While understanding the building blocks of intelligence can be insightful for organizing our knowledge about human and animal cognition, I argue that this approach often lacks the substantiation and concreteness needed for mechanistic understanding—the level at which we must understand intelligence in order to engineer it in artificial systems.

First, core knowledge systems often lack strong empirical support in terms of whether they are, in fact, innate. Abundant studies show that the hypothesized core abilities are present in different species (e.g. primates and humans), early in development, and across cultures (see brief review in Spelke & Kinzler, 2006). However, most of this evidence comes from infants and animals weeks, months, or years after birth, which makes it difficult (if not impossible) to determine whether those abilities are innate or learned. Indeed, there are now many computational existence proofs showing that canonical core abilities—including object perception, face recognition, and navigation abilities—can be learned using generic, domain-general learning mechanisms (e.g., Banino et al., 2018; O'Reilly, Wyatte, & Rohrlich, 2017; Yamins & DiCarlo, 2016). There is also evidence that this learning can happen rapidly. For instance, deep predictive learning is a biologically-inspired framework in which learning involves making predictions about what the senses will report at 100 ms intervals and adapting synaptic weights to improve the prediction accuracy (O'Reilly et al., 2020). Predictive coding is suggested to operate at a rate of ~10 times per second, allowing learning to accumulate rapidly as newborn brains experience the world. Even a few hours of visual experience might be sufficient to develop a "core" understanding of objects and their mechanical interactions. It is therefore plausible that core knowledge systems are learned rapidly using domain-general learning mechanisms, rather than being hardwired into the brain.

Second, the existence of causally important, high-level, and human-interpretable building blocks that might *not* be innate (e.g. as suggested in Lake et al., 2017) is still hard to substantiate. The weakest point of such proposals lies in the very assumption of the primary importance of these capacities for further cognitive development. If the lower-level domain-general learning mechanisms are proposed to underlie the emergence of these interpretable "building blocks", why should the building blocks, rather than the same lower-level adaptation mechanisms, be primarily responsible for driving further achievements? This additional hypothesis of the causal primacy of the learned high-level interpretable building blocks for the development of human cognitive capacities is hard to empirically test without knowing all the details about how developmental programs produce cognitive abilities. I suggest that, so far, there is not enough empirical and computational evidence to assume that the essential causal drivers of cognitive development are the high-level and psychologically-interpretable components.

Finally, hypothesized core systems and skills are difficult to incorporate into machines in practice. Most of the postulated core components are only formulated verbally in high-level psychological terms, which creates a gap between these psychological constructs and formal artificial learning systems. Unfortunately, the high-level psychological constructs of the core knowledge framework often struggle to translate to the lower-level language of modern connectionist AI systems or suggest any formalization at all: often, the core systems theories just postulate the existence of high-level capacities at organisms' birth, without providing any account on what form these capacities can possibly take. What is an appropriate formalization of the innate number sense? How would core knowledge of objects, intuitive physics, or the Language Acquisition Device be instantiated in a neural network? I suggest that the verbally-formulated core systems which do not make clear predictions about the potential architectural or algorithmic instantiations cannot successfully guide modern AI development.

The aforementioned criticisms do not, of course, undermine the general ideas of the importance of innate constraints and biases for the development of learning systems. It is widely appreciated by engineers and proponents of various paradigms in Cognitive Science that all learning systems, including deep neural networks, need strong inductive biases and constraints to learn efficiently. However, these inductive biases are likely to arise from lower-level properties of the learning mechanisms, perceptual and motor systems that drive our adaptation to the environment (see this perspective developed in Elman et al., 1996; Richards et al., 2019). In my view, a focus on these factors, rather than a focus on core knowledge systems per se, provides a more tractable approach for advancing Artificial Intelligence and for building machines that learn like humans and other animals. I devote section 3.4 to a brief discussion of the mechanisms that give rise to language learning constraints.

## 2.2. Neuro-symbolism: symbols are all we need?

A second idea that gained popularity as a cognitively-inspired approach to AI is neuro-symbolism. Symbolic models commonly correspond to largely pre-engineered solutions to tasks which are achieved by hand-coding high-level representations and rules of their allowed transformations. High-level entities (e.g. "circle" and "dog"), their properties (e.g. "red" and "heavy") and relations (e.g. "above" and "causes") have typically served as pre-coded primitives for symbolic models. For the classic symbolic models, the developers needed to conceive of and pre-engineer most of the primitive representations and their operations that are necessary to perform a given task. Eventually, it has been demonstrated that the models grounded in rich environments can often successfully learn intelligent behaviors without any hardwired symbolic representations (Brooks, 1991). Moreover, the common symbolic solutions, conceived without the real-world in mind, typically could not scale to the naturalistic situations. Context-dependent biological minds achieve tractable and robust adaptive behavior by reacting to their rich environments, rather than operating according to the context-free symbolic rules (Dreyfus, 1972). These observations, combined with other criticisms, resulted in a decline of interest in the symbolic AI.

Even though symbolic approaches are no longer the most efficient way to design learning systems, many cognitive scientists have successfully argued that symbolic algorithms should be combined with connectionist architectures to achieve the best, "human-like", learning outcomes (e.g. Marcus & Davis, 2019; Marcus, 2020b; Lamb et al., 2020). In particular, it has been argued that some symbolic operations (such as compositional recombination of tokens) and pre-engineered structured representations should be integrated with the parallel distributed learning, resulting in the hybrid "neuro-symbolic" architectures. Cognitive Science, as proposed, could guide the process of choosing the promising representations, representational formats, and their operations.

The main strength of neuro-symbolic approach is its ability to design systems with expertise. For example, the neuro-symbolic system for learning mathematics does not have to start from scratch and learn how to perceptually tell apart ones from sevens. Numbers, signs, and other potential primitives could be provided as symbolic inputs to the system, allowing it to learn on these higher-level representations. Thus, this approach can lead to more tractable and efficient learning for many applications. Top-down engineering of the structured representations, however, comes with its disadvantages. It is extremely hard to design the representations that are compressed enough to facilitate the learning process and which also retain all information that is useful for solving a given task. It can be illustrated with the aforementioned example of math learning: the elimination of all perceptual information about numbers would exempt the model from learning to process perceptual inputs, but it would also preclude it from noticing the similarity between 10, 100, 1000, and so on. The more complex the learning problem and the data are, the harder it becomes to choose efficient high-level representations and operations.

The reliance of neuro-symbolic modeling on scientists' insights could be particularly misleading if the goal is to design systems with human-like learning abilities. It has been argued that scientists' intuitions are very limited and could not capture the iterative nature of human learning which deals with the rich multimodal training data in multitask conditions over the years of development (e.g. Brooks, 1991; McClelland et al., 2010; Chemero, 2011; Hasson, Nastase, & Goldstein, 2020). The artificial neural network and robotic simulations demonstrate that high-level pre-engineered solutions are typically very far from the solutions that learning systems discover by themselves, even for the simplest learning problems. It turns out that there are multiple shortcuts, available in the task-specific structure of complex environmental data, motor and perceptual resources of the agents, that allow end-to-end learning systems, such as humans, animals, and Deep Neural Networks, to solve cognitive problems without coming up with the expected structured representations (Hasson, Nastase, & Goldstein, 2020; Geirhos et al., 2020). For example, the artificial agents that learn to categorize a relation between sizes of two consecutively presented objects sometimes end up encoding the object size with their body position instead of memorizing it "internally" (i.e. through the neural network weights or activations) (Williams, Beer, & Gasser, 2008).

Often, the solutions that are needed to capture useful patterns in rich datasets are much more or much less nuanced than the ones anticipated by scientists. The quality of the learnt shortcuts, however, is typically evaluated by how well the model performs on the intended generalization domain (Geirhos et al., 2020). Thus, whenever the training domain is representative of the target application domain (see section 3 for the discussion of essential properties of naturalistic communication contexts), the learned shortcuts, no matter how interpretable or anticipated they are, can be very adaptive (Hasson, Nastase, & Goldstein, 2020). The observations of unexpected efficient shortcuts that the artificial and natural minds often discover for achieving their goals question the validity of intuitive and interpretable pre-engineered solutions and representations that cognitive scientists rely on when they adopt the symbolic approach. While the symbolic or hybrid (neuro-symbolic) solutions might work well for the artificial tasks, they might not be as efficient for solving real-world problems. I suggest that relying too heavily on human intuitions about efficient representational structures in AI might not be the best path towards building machines that develop human-like linguistic intelligence.

Combining symbolic and neural approaches has recently resulted in many success-stories from different domains. However, at least part of the success of these systems can be attributed to the opportunity to pre-engineer the representational structure to fit the structure of the training and evaluation tasks. For example, in a recent study, Adhikari et al. combined structured graph-based representations and end-to-end learning to solve a large set of text-games (Adhikari et al., 2020). This model outperformed the previous approaches by 24%, which is a truly impressive result! However, a closer examination reveals that the states of the used text-games adhere to the same dynamic graph-structure (entities, conditions, and their relations) that was used in the model. Thus, the representational structure of the model fits the structure of specific text-game benchmarks, allowing it to efficiently uncover the necessary relationships to solve the task. I suggest that the generative structure of real-world observations is a black-box that is far more complex than the tasks that are used to train and test neuro-symbolic models. In fact, learning the task-relevant structure of different environments is one of the essential skills that autonomous and adaptive biological systems need to master.

In this paper, my goal is to draw attention to the alternative, "bottom-up" approaches in Cognitive Science, which focus on the properties of rich naturalistic learning conditions and on the power of generic, "low-level", end-to-end learning mechanisms that drive the development of human cognitive capacities.

## 2.3. GROUNDED LANGUAGE ACQUISITION

Extended, Embodied, Embedded, and Enactive (4E) proposals approach the questions of cognitive development from a different angle. The proponents of these paradigms focus on the properties of natural learning settings, which are indispensable from the results of human and animal cognitive development. A common methodological basis of the 4E research is bottom-up theorizing and modeling: instead of relying on human intuition about how agents are able to learn and execute new behaviors, the followers of this approach empirically test, simulate, and analyze learning in naturalistic conditions to see what is "already provided" by these settings alone, and how a learner achieves the rest (Gibson, 1979; Brooks, 1991; McClelland et al., 2010; Chemero, 2011; Clark, 1998; Hasson, Nastase, & Goldstein 2020). Often, this approach leads to discoveries that the supposedly complex intelligent behavior can arise from surprisingly straightforward interactions of an agent and its environment (for review, see Cangelosi et al., 2015; also, see section 3.1.2. for further discussion). Therefore, my further analysis of 4E components for building linguistic intelligence mainly focuses on the necessary aspects of *simulation design*, rather than on particular architectures or skills pertinent to a learner.

4E perspectives criticize the dominant AI research as being unable to capture properties of the learning conditions that heavily guide the development of animal and human intelligence. Most of these claims do not contradict the Parallel Distributed Processing paradigm for modeling learning systems (McClelland et al., 1986), but rather shift the focus from searching for more powerful architectures and bigger datasets to the crucial aspects of learning settings, which are underrepresented among AI research practices. In the following sections, I articulate the "core" aspects of real-world learning conditions and review evidence supporting their role in language development. I also show how some of these principles have been applied in machine learning simulations. Finally, I propose a list of concrete recommendations for incorporating these "grounded" insights into AI practice.

This review is not first to propose a grounded perspective on AI: prominent scientists in the fields of Embodied Cognition and Robotics have argued for similar changes in AI research (Neisser, 1993; Brooks, 1991; Steels & Brooks, 1995; Smith & Gasser, 2005; Pezzulo et al., 2011, 2013). Moreover, the small yet prolific fields of Cognitive and Developmental Robotics have instantiated these "grounded" principles in AI. Recently, the machine learning community has seen a rise of proposals to "ground" language acquisition (McClelland et al., 2020; Baroni, 2020; Tamari et al., 2020; Bisk et al., 2020; Gauthier & Mordatch, 2016; Kiela et al., 2016). These proposals indicate a growing interest in 4E ideas for building communicative AI. However, these proposals lack a comprehensive analysis of different aspects of "language grounding", which has been an area of extensive research in Cognitive Science for several decades. This review complements these earlier contributions by putting the grounded perspective in the context of other directions of Cognitively-Inspired AI, articulating a set of distinguishable aspects of grounded language learning, reviewing empirical evidence for the role of these different aspects in language acquisition and processing, and formulating a list of concrete principles to accelerate the application of these insights in the common practice of AI research. The field of AI needs clear understanding of what it actually means "to ground language" – which, I hope, will be facilitated by this paper.

## 3. GROUNDED LANGUAGE LEARNING IN HUMANS AND AI

Which aspects of natural learning help humans acquire languages? In this section, I will try to divide the elusive "language grounding" idea into concrete, distinguishable components and evaluate their contribution to human language abilities. I integrate evidence from computational simulations and empirical studies of human language learning and communication. This analysis then motivates a set of concrete principles for future AI research, where each component can be applied independently from another, according to whether a researcher is convinced by the presented evidence.

## 3.1. LANGUAGE IS ACQUIRED IN RICH MULTI-MODAL CONTEXTS

Human language acquisition, comprehension, and production are often situated in multimodal perceptual contexts. Rich multimodal environments heavily guide these communication processes, making the ambiguous clear, underspecified – obvious, individual – shared, and computationally-complex – easily solvable.

### 3.1.1. MULTIMODAL COMMUNICATION IS REDUNDANT, ROBUST, AND ADAPTIVE

Multimodal contextual information helps to disambiguate noisy and underspecified utterances, assess communicative intentions, and track partner's understanding. Communicative information is transmitted through many different channels, rather than just one, making it redundant and more robust to noise and variation. Human speech production in multimodal contexts, in turn, is adapted to the redundant, non-verbally and contextually augmented transmission, which makes it more likely to omit information that is present in the scene or conveyed by communicative actions. Moreover, multimodal transmission opens up rich potential for coordinating activity, which leverages understanding among individuals with different prior experiences and capabilities.

Humans integrate multiple modalities to disambiguate linguistic information on the word, sentence, and discourse levels (e.g. Henderson & Ferreira, 2013; Tanenhaus & Trueswell, 2006; Eberhard et al., 1995; Spivey et al., 2002). For instance, during spoken-language comprehension, listeners' spontaneous eye-movements actively track potential referents, taking into account visual contextual information when interpreting speech (e.g. Spivey et al., 2002). Other results suggest that adults integrate this multimodal information during spoken-language comprehension proactively, combining the visual and lexical information to anticipate the lexical information that has not yet appeared (e.g. Kamide, Altmann, & Haywood, 2003; Altmann & Kamide, 2007). For example, when human adults hear a sentence that starts with "the man has drunk", they are more likely to look at the empty instead of a full glass than when they hear a sentence that starts with "the man will drink". Child language acquisition and comprehension also rely on contextual guidance. After three years of age, children begin showing some predictive patterns of multimodal cue integration during speech processing that were found in adults (e.g. Borovsky, Elman, & Fernald, 2012; Nation, Marshall, & Altmann, 2003; Snedeker & Trueswell, 2004). Moreover, a recent analysis of a massive longitudinal dataset of audio and video recordings of one child for 3 years after birth found that the distinctiveness of the visual contexts of a word's usage is more predictive of its early production than its frequency in caregiver's speech (Roy et al., 2015). Thus, visual, motor, and auditory sources of information all contribute to efficient human language learning and communication.

Spontaneous movements during naturalistic communicative interactions help people with different background knowledge convey and comprehend meaning. The coordinated movement patterns (e.g. eye movements, pointing, and gesturing) have been suggested to actively manipulate the common ground by participants in dialogues (e.g. Clark & Brennan, 1991; Clark, 1996; Garrod & Pickering, 2009). For example, when the eye movements of both the speaker and listener continuously indicate their current understanding, misunderstandings can be easily tracked and fixed. A controlled experiment on speech production and comprehension with the shared visual context demonstrated that the strength in which a speaker and listener's eye movements are coupled predicts the success of listener's understanding (Richardson & Dale, 2005). A recent meta-analysis corroborates this idea by showing that spontaneous gesturing enhances listener's understanding and often provides some novel, non-redundant communicative information (Hostetter, 2011). Thus, a great deal of information in naturalistic communicative interactions is provided non-verbally: speakers continuously display their intended meaning, while listeners respond by signaling their current understanding.

Similar multimodal coordination mechanisms guide early language learning. Children benefit from gesture that accompanies their partner's speech even more than adults (for meta-analysis, see Hostetter, 2011). On the other side, gesture spontaneously appears in young infants, and it was suggested to complement verbal information and indicate infant's understanding (e.g. Goldin-Meadow, Alibali, & Church, 1993; Iverson, 2010; Colonnesi et al., 2010).

Thus, the "multi-channel" nature of human communication underlies successful communicative interactions and linguistic inferences in many naturalistic situations (Fig.1). Human speech, in turn, is adapted to the redundant transmission that characterizes naturalistic communicative contexts, allowing much more variation and noise than machines, which operate on amodal linguistic signals, can reliably process.

In contrast, most of the AI models used in language tasks are trained exclusively on amodal textual excerpts. For example, the state-of-the-art models GPT-2 (Radford et al., 2019) and GPT-3 (Brown et al., 2020) were trained on 40GB and 45TB of textual data from the Internet, respectively. These models do not benefit from the redundancy of information that is transmitted through multiple channels. More importantly, these kinds of models do not operate over information sources that are often *necessary* to account for the variability of human speech in multimodal contexts, and hence could be unreliable when applied in multimodal communicative situations.

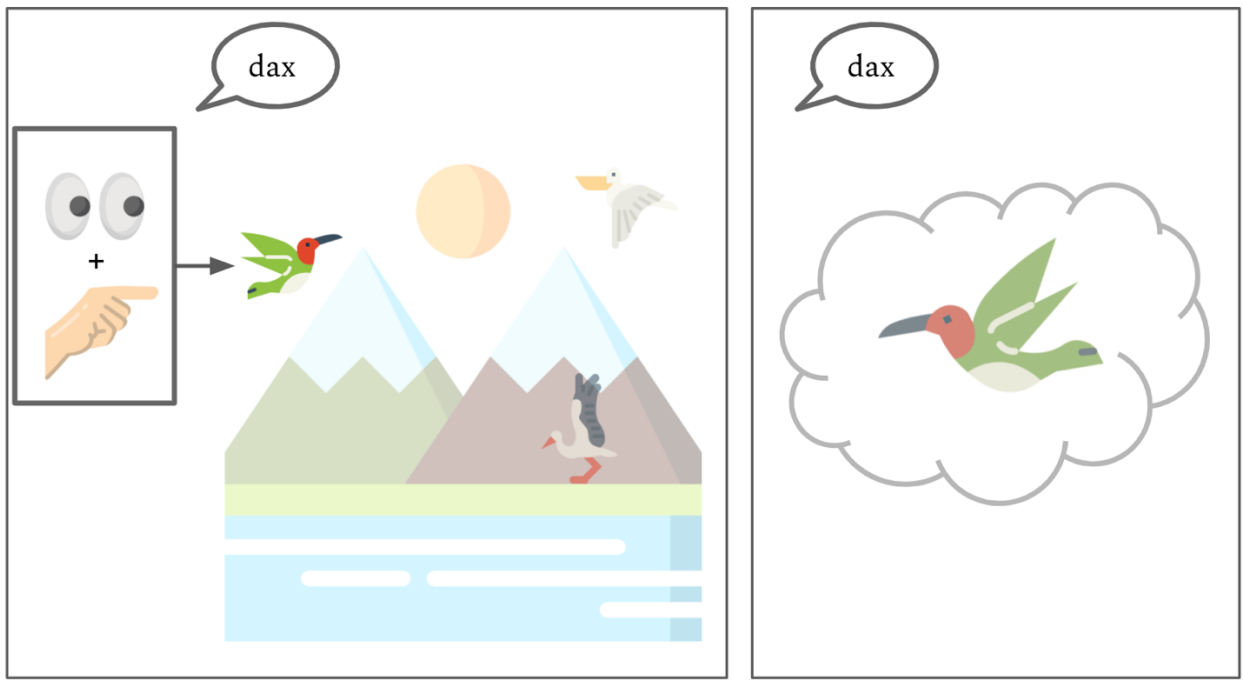

Figure 1. Cues in the naturalistic language learning environments support efficient meaning disambiguation (left). The acquired concepts can then be perceptually simulated during language understanding (right).

The figure uses materials from flaticon.com

### 3.1.2. DIRECT-FIT TO TASK-RELEVANT CONTEXTUAL INFORMATION

The "disambiguating" role of contextual information for language understanding likely understates its importance: the contextual information is often a "primary", rather than "secondary", source of information that is used for structured linguistic inferences. Both empirical studies (partially reviewed in the previous section) and computational analyses demonstrate that human communicative environments provide much more task-relevant information than is often considered. However, for the cases of "higher-level" cognitive abilities, such as language, the role of seemingly "irrelevant", but in fact essential for human abilities, cues in the environment is under-studied and, potentially, under-appreciated.

Robotic simulations demonstrate that artificial agents that can interact with their environment often solve so-called "cognitive" tasks by finding simple "direct" perception-action shortcuts. For example, agents confronted with the problem of shape categorization evolve a solution that involves actively moving the body so that shape-diagnostic information can be obtained directly from the sensors, rather than by building explicit shape-based internal representations (Beer, 2003). This and other simulations (for a review, see Cangelosi et al., 2015) demonstrate that models focused entirely on building and manipulating disembodied representations to solve cognitive problems might be missing a variety of other solutions that animals use for solving naturalistic tasks in rich environments. The natural solutions are often quite different from the clever, internalized rules and structured representations that human engineers come up with to solve the same problems (e.g. Brooks, 1991; Hasson, Nastase, & Goldstein, 2020).

Even though much work in analyzing and quantitatively evaluating language-related structure available in natural communicative environments remains to be done, some results from other areas suggest the fruitfulness of this approach. First of all, it has been convincingly demonstrated that "direct" perception can underlie seemingly "high-level" cognitive inferences, such as determining agents' intentions (e.g. by motion cues in Barrett et al., 2005). Secondly, rich communicative information is often directly available in gaze direction, gesture, pointing, the visual structure of the scene, and many other sources (e.g. Tomasello, 1992; Baldwin, 1995;

Gogate, Walker-Andrews, & Bahrick, 2001; also see the studies reviewed in the previous subsection).

The contextual effects, described in these two subsections, demonstrate the richness of information available to individuals participating in naturalistic communicative interactions. Most of the aforementioned studies, however, are conducted in visual and auditory domains. Contextual information from other modalities, such as touch, can also be integrated in this process. Congenitally blind individuals might consistently use the structure available from non-visual sources to disambiguate linguistic information. Some evidence suggests, however, that blind children acquire language slower than sighted (Andersen, Dunlea, & Kekelis, 1993) and deaf children (Preisler, 1995), indicating the importance of vision for human language learning. On the other hand, the word co-occurrence structure captures many aspects of the world (Landauer & Dumais, 1997), such as visual similarities between objects (e.g. animals) and the associations between visual and non-visual features (e.g. color-adjective associations: "red is aggressive", "blue is cold"), allowing blind people to gradually pick up some information that is unavailable to them by sense (Lewis, Zettersten, & Lupyan, 2019; van Paridon, Liu, & Lupyan, 2021).

In sum, when children learn language, they leverage the rich, natural, and interactive communicative experiences provided in the world. These experiences include the child's early language environments and conversations with friends and family members. Of course, once a person has acquired language, they can easily comprehend and transmit meaning in other contexts, such as listening to a podcast or talking on the phone. This fact indicates that rich multimodal communication settings are not necessary for understanding the acquired language, but these settings may be necessary for the initial development of human language capacities, which is always embedded in rich environments. As we will see in the next subsection, there is a growing body of evidence suggesting that *interiorization* of perceptual experiences might take place throughout development, enabling us to *simulate* multimodal perceptual contexts in order to understand amodal and passive language (e.g., when reading a book or listening to a podcast). Rich perceptual experiences accompanying communicative interactions during language acquisition are proposed to be a necessary foundation for such change.

### 3.1.3. RICH PERCEPTUAL EXPERIENCE GROUNDS PERCEPTUAL SIMULATIONS FOR LANGUAGE UNDERSTANDING

Many psychological and linguistic theories hypothesize that humans comprehend language by perceptually *simulating* the respective scenes (Fig.1). This proposal has been developed in L. Barsalou's Theory of Perceptual Symbols (Barsalou, 1999) and in other views arguing for modal simulations underlying language understanding (e.g. Zwaan & Radvansky, 1998; Glenberg & Robertson, 1999; Grush, 2004; Gallese, 2008; Zwaan, 2016). The Perceptual Symbols theory integrates symbolic processing with perception and action to explain many "high-level" cognitive phenomena, including categorization and language. The main idea of perceptual simulations is that symbolic operations, commonly considered to be a crucial component of human thought, operate on modal (sensorimotor), rather than amodal, representations. According to this proposal, the perceptual experiences (e.g. seeing a moving car) are stored in frames consisting of distinguishable components (such as a wheel, window, and frame of a car). The memories of the components are organized around similar frames and can be recombined during language comprehension. In a similar way, motor simulations are suggested to play a central role in language acquisition and processing (e.g. Glenberg & Robertson, 1999; Glenberg & Gallese, 2012). In particular, it is argued that many utterances (e.g. verbal instructions, descriptions of actions) can only be understood if they can be integrated in a simulation of a sensible goal-directed action. Even though most perceptual simulation models of language comprehension offer a conciliatory perspective between symbolic and embodied approaches (Zwaan, 2014), they emphasize an essential role of perceptual and motor experiences as primary units of language understanding.

Thus, language comprehension has been argued to involve reconstruction of modal events from previous sensorimotor experiences. To infer meaning of a new utterance, a person is suggested to simulate the perceptual and motor aspects of the respective scene. Importantly, recombination of the word-part meanings happens in a way that "makes sense" given one's previous experiences with the world or affordances of the objects involved. For example, "blue bus with yellow dots" can be understood as a bus with an unnaturally colored body, while the wheels and windows maintain their typical colors.

The idea that language comprehension involves re-activation of perceptual, motor, and emotional events is consistent with empirical work in psychology and neuroscience (for reviews, see Pulvermüller, 2005; Fischer & Zwaan, 2008; Binder & Desai, 2011; but also Caramazza et al., 2014). One such study found that the words referring to similar motor actions (e.g. playing a piano and typing on a computer) prime each other in a lexical-decision task, even after controlling for the visual similarity between these actions, suggesting that the motor representation of an action is re-activated during respective word processing (Myung, Blumstein, & Sedivy, 2006). The other study demonstrated that listening or reading about the actions that are associated with different parts of the body (e.g. "kick", "pick", "lick") leads to rapid (130-170ms) activations in the respective parts of sensorimotor brain areas, which somatotopically encode leg, hand, and mouth movements (Pulvermüller, Shtyrov, & Ilmoniemi, 2005). The follow-up TMS experiment has completed this evidence by showing that controlled inhibition of the topographic brain regions associated with action production in different parts of the body (leg or arm) leads to significantly slower response times in the word recognition test for the words that were associated with the body part whose motor localization was actively inhibited (Pulvermüller et al., 2005). In a recent computational study, De Deyne and colleagues (2021) compared unimodal (linguistic) and multimodal (linguistic + visual + affective) semantic models. Their analysis confirmed that the models that contain information from multiple modalities are better predictors of human semantic similarity judgements than their counterparts trained only on the linguistic information. Together, these results demonstrate that the multimodal information is reliably re-activated by words and that this information constitutes human semantic knowledge.

Tamari et al. (2020) have recently argued for the promise of applying the perceptual simulation ideas for developing communicative AI. Similarly, Hamrick (2019) suggested that *model-based* Reinforcement Learning algorithms essentially employ "simulation-based thought" to learn effective policies. These models solve different tasks (e.g. playing ATARI games) by learning to predict the future states of the environment and make decisions on the basis of these expectations. The "internal models" of the environment that are used to make the state predictions, in turn, can be used to simulate the outcomes of different actions in a given or hypothetical situation, or to construct a respective perceptual scene for an utterance.

## 3.2. LANGUAGE IS ACQUIRED BY AUTONOMOUS AGENTS DETERMINING THEIR OWN EXPERIENCES

Active embodied exploration in the environment lets human language learners determine their learning data through efficient information-seeking mechanisms, tailored at one's own current progress. Because human children can choose how to explore the world at any moment, language learning tends to proceed hierarchically: from simpler skills and patterns, more salient objects, and more predictable aspects of the world to more difficult, allowing cognitive abilities to gradually grow on top of each other (Vygotsky, 1964; Thelen & Smith, 1996; Di Paolo, Cuffari, & De Jaegher, 2018).

### 3.2.1. DEVELOPING HUMANS CHOOSE THEIR OWN CURRICULUM FOR LANGUAGE LEARNING

Children and adults exploring in rich environments determine their own training data and the sequence of its presentation (Fig.2). This opens multiple possibilities for structuring learning with respect to an individual's state of knowledge, needs, and cognitive characteristics at the moment. These possibilities include spending more time actively exploring new rather than old items, obtaining as many views of an object as needed to discriminate that object from already learned ones, asking questions when something is unclear, generating caricature examples to get feedback on the category's diagnostic properties, as well as taking into account the characteristics of one's own memory, motor and perceptual systems when choosing the learning strategies (Smith et al., 2018; Gureckis & Markant, 2012).

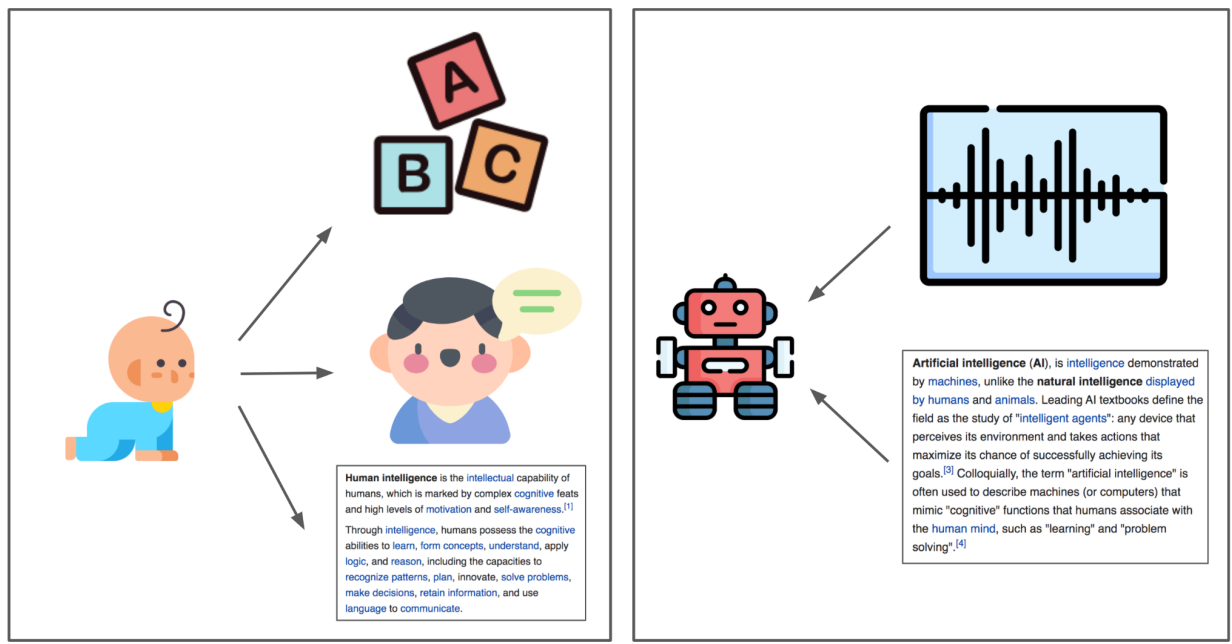

Figure 2. Active and passive learning. Humans can choose what to learn (left), while machines passively receive the training examples (right).

The figure uses materials from the following sources:

Flaticon.com

Wikipedia contributors. (2021, February 12). Artificial intelligence. In *Wikipedia, The Free Encyclopedia*. Retrieved 21:17, February 14, 2021, from https://en.wikipedia.org/w/index.php?title=Artificial_intelligence&oldid=1006429124

Wikipedia contributors. (2021, February 9). Human intelligence. In *Wikipedia, The Free Encyclopedia*. Retrieved 21:35, February 14, 2021, from https://en.wikipedia.org/w/index.php?title=Human_intelligence&oldid=1005838764

Strong evidence for the importance of self-generated data for early word learning comes from studies using head-mounted cameras in naturalistic learning conditions. In real life, children actively manipulate objects, turning them around, moving them closer and farther away from their eyes, and exploring them under different physical conditions. By moving and densely sampling views of objects, infants can acquire the visual training data needed to build object representations that are generalizable across contexts and viewpoints (Smith et al., 2018; Orhan, Gupta, & Lake, 2020). This self-generated visual variability has been found to predict the success of learning the object's name (Slone, Smith, & Yu, 2019). Studies with adults demonstrate the advantages of active learning as well (for review, see Gureckis & Markant, 2012). For example, humans learn new categories more efficiently when they can actively query new examples (Markant & Gureckis, 2010). These results suggest that *active* language learning is often more efficient than *passive* learning because active learning allows agents to choose the most informative data for their knowledge acquisition.

Recent work with embodied artificial agents provides a convincing demonstration of the central role of active egocentric perception in language learning (Hill et al., 2019). The artificial agents were trained in 2D and 3D worlds, and then tested on their ability to understand new combinations of already learned concepts. When the agents were trained with a bounded (egocentric) reference frame, the agents developed enhanced abilities to generalize their semantic knowledge to novel situations. These agents that could move in the environment, choosing their own perceptual inputs, performed better at compositional generalization than the "passive agents", who received the same set of training data but could not actively select a bounded subset of inputs at each moment. This study highlights the importance of embodiment and egocentric active learning for systematic compositional generalization of linguistic knowledge.

Vast and open-ended information spaces of natural learning require powerful exploration mechanisms to guide search for useful data. Various information-, uncertainty-, and novelty-based exploration mechanisms have been hypothesized to generate structured data for early learning in human babies. The empirical data show that children generate observations by actively searching for information in the environment and keeping track of their learning returns. In particular, infants allocate their visual (Kidd, Piantadosi, & Aslin, 2012) and auditory (Kidd, Piantadosi, & Aslin, 2014) attention to the stimuli of "optimal" complexity (neither too predictable nor too surprising for them), which supports learning that is adapted to the current state of their development. Evidence for preferential attention to more "learnable" linguistic patterns has also been found in 17-month-old infants habituating to words generated with rules of different complexity (Gerken, Balcomb, & Minton, 2011). Preliminary results from Zettersten's dissertation (2020) suggest that children prefer to request additional demonstrations for infrequent or ambiguous word-object pairs, but these preferences might only emerge around 3 and 5 years of age (respectively) (see also Zettersten & Saffran, in press). The middle schoolers and adults starting to learn the words of their second language also preferentially allocate their studying time to medium-difficulty examples, which leads to the best results in this learning problem (Metcalfe, 2002; Metcalfe & Kornell, 2003). Thus, familiar or too difficult objects, words, or aspects of the environment receive less human attention and exploration time, while more "learnable" items draw more attention. This adaptive allocation of learning time and resources gradually changes with acquisition, making a path for new concepts and skills.

Overall, the existence of domain-general curiosity-driven exploration mechanisms in humans is supported by a large body of empirical work in neuroscience and psychology (for review, see Gottlieb et al., 2013; Kidd & Hayden, 2015; Schulz, 2012). These curiosity-driven exploration mechanisms are hypothesized to guide gradual development of "high-level" cognitive capacities, such as tool usage and language, and might be tightly connected to the developmental programs that are optimized by evolutionary search (this perspective is outlined in Oudeyer & Smith, 2016).

It would be hard to overestimate the role of individualized epistemic activity in organizing human language learning. Adaptive exploration mechanisms pave a unique path for language development in children with different sensorimotor and cognitive characteristics, developmental states, and learning conditions (Thelen & Smith, 1996; Oudeyer, 2006; Di Paolo, Cuffari, & De Jaegher, 2018).

Adaptive exploration in human babies has inspired simulations of language development in the field of Cognitive Robotics (for review, see Oudeyer, Kachergis, & Schueller, 2019). For example, Oudeyer and Kaplan (2006) conducted experiments of exploratory learning in an autonomous curiosity-driven robot. The robot's behavior resembled the developmental stages of human babies. The robot first engaged in exploration of its body capacities (e.g., moving individual actuators and observing the outcomes) until they became predictable. Then, the robot started to explore affordances of the objects, by trying to bite or bash different toys and discovering the consequences. After learning about the objects in the environment, the robot started to play with the other, "speaking" robot. Notably, the speaking robot had its own operation rules and, thus, was the most difficult to predictably control. The curious robot even spontaneously started to babble in order to provoke the other robot's responses. This simulation suggests that some developmental transitions found in human babies' behavior may naturally arise from autonomous exploration driven by domain-general information-seeking objectives ("curiosity"). Such robotic experiments serve as proofs-of-concept that the domain-general curiosity-driven exploration can allow agents grounded in rich environments to independently achieve goals of increasing complexity, such as locomotion, object recognition, and communication.

The active learning strategies are particularly useful in cases when supervising feedback is rare or expensive, but the unsupervised exploration is cheap (Settles, 2012). This perfectly characterizes the problem of learning languages in multimodal social environments, where the agent can endlessly explore objects and actions on their own, but requesting labels of these entities either takes more time and is not always available. In this case, the learner is most effective when they can make the best use of the limited supervising feedback (e.g. labels) by requesting the information that has the highest expectancy of being useful given learner's knowledge and the structure of the environment. The benefits of active learning, however, have yet to be fully integrated in the development of artificial systems for language learning (Fig.2). Most of the state-of-the art models for language processing either sample their observations randomly or choose the observations on the basis of pre-engineered curriculum rules (e.g. see Devlin et al., 2019). I suggest that focusing on the active learning models could bring us closer to building the agents that could effectively learn in rich communicative environments.

## 3.3. Language Arises and is Maintained in Interaction of Bodies, Brains and Environments

Humans acquire languages in adaptive and controllable environments. These environments are full of other communicative agents, who adapt their signals to the learner's current state of knowledge and afford "requests" for help and clarification (e.g. by asking a question, looking confused, not responding to verbal and non-verbal interactions). These and many other learner-adaptive properties of human environments dramatically reorganize and scaffold human language learning (Di Paolo, Cuffari, & De Jaegher, 2018; Hasson et al., 2012; Gogate, Walker-Andrews, & Bahrick, 2001; Steels, 2000).

### 3.3.1. Languages are acquired through adaptive social interactions

Acquired communicative systems in animal and human groups develop through biologically grounded social events. The adaptive properties of communicative social interactions shape learning mechanisms that are needed for successful language acquisition, as both the feedback and data samples are often adapted to the individual's state of knowledge, goals, as well as their sensorimotor and cognitive constraints.

Social interactions endow young human language learners with adaptive data and feedback. In fact, humans demonstrate high sensitivity to adaptive multimodal social feedback and communicative history from the earliest stages of development (Goldstein & Schwade, 2010). For example, infants' prelinguistic vocalizations are affected by interactive social feedback from their caregivers (e.g. Goldstein, King, & West, 2003; Goldstein & Schwade, 2008), who provide differential responses depending on the speech-like quality of infant's signals (Gros-Louis et al., 2006). At first, caregivers simplify phonetic and lexical structure of their vocalizations with respect to the vocal maturity and phonetic level of the learners (e.g. Kuhl et al., 1997; Elmlinger,

Schwade, & Goldstein, 2019). Moreover, caregivers produce shorter utterances and more repetitions when they talk to younger infants, gradually changing to producing less repetitive, more syntactically rich and phonetically diverse speech as the child grows (for review, see Soderstrom, 2007). On the prosodic level (e.g. reflecting intonation and rhythm), however, infant-directed speech is less predictable than adult-directed speech, and thus potentially more attractive to curiosity-driven babies (Räsänen, Kakouros, & Soderstrom, 2018; MacDonald et al., 2020). Caregivers also promptly correct children's incorrect utterances, providing immediate supervising feedback on phonological, morphological, syntactic, and lexical levels. There is evidence that this kind of language supervision declines as the child grows (Chouinard & Clark, 2003). Thus, it has been argued that caregivers provide adaptive vocalizations and feedback that flexibly adjust to the learner's development, thereby scaffolding early language learning (e.g. Gogate, Walker-Andrews, & Bahrick, 2001; Hasson et al., 2012; Di Paolo, Cuffari, & De Jaegher, 2018).

Learner-specific data and feedback have been suggested to influence language acquisition in sensory impaired children. For example, there is evidence that the caregivers of blind children are less likely to talk about people and objects in the environment, unless the child is actively interacting with them through touch. They also use more descriptions of the ongoing child's actions than caregivers of the sighted children, adapting to what is likely to be present in the child's experiences at a given moment (e.g. Andersen, Dunlea, & Kekelis, 1993). Likewise, deaf mothers of children adapt their gestures to attract deaf infant's attention using various strategies, i.e., gesturing in the child's field of view (so the child does not need to move her gaze), as well as applying tactile signaling to indicate communicative intention, which decreases in frequency as the child's communicative capacities improve (Waxman & Spencer, 1997; Spencer & Harris, 2006). Thus, the flexibility of human communication and language teaching may support language learning in children with very different developmental starting points (Thelen & Smith, 1996).

Adaptive nature of human social interactions underlies successful conversations (Pickering & Garrod, 2004; Garrod & Pickering, 2004; Dale et al., 2013). For instance, human auditory perception quickly adapts to particular partner-specific speech, so that the words pronounced by the familiar speakers are more intelligible (Nygaard & Pisoni, 1998), while the speech production of participants in a conversation couples in terms of the speed, accent, phrases, and the referent names that are used (e.g. Giles, 1973; Giles et al., 1987; Pickering & Garrod, 2004). Similar language coupling has been shown between children and their caregivers: for example, they dynamically "coordinate" their speech on syntactic level, so that both a child and a caregiver tend to use the syntactic sequences (e.g. verb-pronoun) that match the ones being heard (Dale & Spivey, 2006). Thus, our everyday communicative interactions are facilitated by the mechanisms that adapt our language perception and production to every new partner.

AI systems, on the contrary, rarely learn to communicate from humans or from other artificial agents. Even when humans are involved in evaluating the communicative attempts of an artificial agent, these evaluations are rarely individualized, adaptive, or shaped by the supervisor's intuitions about the machine's current state of knowledge and learning limitations. These properties of natural language learning are even absent in multi-agent communicative learning simulations: agents do not learn to "teach" each other, and the communicative feedback typically comes uniquely from the environment (e.g. Lazaridou et al., 2018; Foerster et al., 2016; Mordatch & Abbeel, 2018).

Some preliminary developments in social robotics, however, demonstrate the promise of learning to communicate by interacting with humans (for a brief overview, see Broz et al., 2014). For example, Lohan et al. (2012a) developed a robotic system that adaptively responds to human teaching by following the tutor's gaze and pointing in response to the tutor's object demonstrations. The preliminary results showed that the adaptive social feedback from the robot provokes particular type of teaching from human tutors: for example, when demonstrating the patterns on objects to the robot, human tutors used at least two times more words teaching the interactive robot in comparison to teaching the robot that does not follow their gaze and points to the objects at random moments (Lohan et al., 2012b). I suggest that the benefits of social language learning in AI should be further explored for building machines that are capable of efficient communicative interactions with human partners.

### 3.3.2. LANGUAGE CAPABILITIES CAN BE PARTIALLY OUTSOURCED TO CONTROLLABLE ENVIRONMENTS

We live in highly controllable environments. Thus, we have multiple opportunities to outsource parts of our cognitive processing to the world, without having to perform all cognitive operations internally (Clark, 1998; 1999; 2013). For example, we can easily do multiplication of large numbers by writing simple calculations on paper, while most of us find it difficult to perform these operations uniquely within our heads. Our notebooks operate as our external memory, and our laptops with the Internet replace the need to remember all useful information; this information can always be retrieved once we know how to use a search engine. In this context, again, our brains are better viewed as effective manipulators of external tools and artifacts, rather than as cognitive processors uniquely responsible for all the computations that drive our behavior. Thus, human-like learning involves efficient interaction with tools and devices, with no necessity to replicate their capacities.

Potential externalized strategies that children might employ for efficient language learning have been understudied. However, after mastering language, a big part of our linguistic knowledge can potentially exist "outside" of our body, always available for retrieval. For example, we can look up the facts that we are not sure about, check the grammatical rules, and request more examples of word usage in both linguistic or multimodal contexts from gadgets or from other people. These resources are available for us almost all the time, and therefore, should not be dismissed in their potential to restructure the linguistic knowledge that we assume modern humans possess. In similar ways, these external resources might be important for second language acquisition. In the absence of a constant "linguistic environment", the ability to find and use learning resources allows us to ask questions and request feedback to adjust our language skills to the ever-changing linguistic climate. We can listen to the pronunciation of phonemes, words, and intonations of native speakers, try to perceptually differentiate them, imitate and get feedback from language-learning applications, watch videos with contextualized native speech and subtitles at comfortable speed, look for translations of unknown words and phrases, choosing learning materials which fit our current knowledge and limitations. When human mentors and conversational partners are not always available, the books and Internet become powerful replacements, allowing learners to search for important information in a variety of new ways.

Modern AI work on language, however, is predominantly focused on training machines in a "learning vacuum". AI systems are typically trained to produce amodal responses, matching given inputs to desired outputs. In these passive pattern-matching problems, there is no history of interactions between an agent and its environment. Often, there is no environment in the first place. Such conditions put machines at a disadvantage: unlike humans, communicative AI systems don't operate in environments that could potentially *extend* what they can do by providing ways to use and exploit external resources for their learning.

In contrast, when artificial learning systems can continuously interact with their environment, they often spontaneously restructure the surroundings to support their goals. In these cases, the agents learn to manipulate objects or other agents to get what they need. For instance, in a recent multi-agent reinforcement learning simulation, agents learned to play competitive games in virtual environments with several controllable elements. In the hide-and-seek game, the hiding team learned to use cubic objects to lock themselves in confined spaces, so the seekers could not reach them and win. Then, the seekers learned to use the objects as steps to jump through the walls to counteract the hiders' strategy (Baker et al., 2019). These tool use strategies were not pre-programmed and were not directly related to rewards: the agents could benefit from tools only when they appropriately use them, often in ways that the engineers did not expect at the beginning of simulations. These experiments demonstrate that artificial agents can spontaneously start using tools when the learning environments contain elements that the agents can control.

Some recent developments in robotics take inspiration from human ability to leverage external resources to support lifelong learning. For example, Mancini et al. (2019) describe the "open-world" image classification system that successfully deals with the situations when the new types of objects continue to appear after the initial training. When the system recognizes a novel class of objects, it tries to mine similar images with their labels on the internet to learn about this new class. The preliminary results demonstrate that this system is able to reach higher than baseline performance on the novel classes and that it can be successfully deployed in physical robots. I suggest that controllable environments and external resources (e.g. internet) should be considered for building artificial agents that could restructure their surroundings to support their communicative goals and quickly update their linguistic knowledge to the ever changing environment.

### 3.3.3. Developing brains, developing bodies, and developing environments

The course of sensorimotor and physiological development dynamically determines what kind of sensory data the baby can acquire, what actions they can perform, what motor skills are possible to learn, and what prior background and skillset they have in their disposition (Thelen & Smith, 1996; Iverson, 2010).

One potentially powerful idea is that motor development places strong dynamic constraints on the kinds of problems that an agent can learn. By gradually increasing the action and perceptual spaces of the agent, motor development adaptively constrains the possible child-world interactions, allowing knowledge to emerge gradually over time (Thelen & Smith, 1996; Iverson, 2010). In humans, early language learning involves mastering simplistic, exaggerated utterances, simple objects, and coarse-grain motor responses. Later, as the space of available motor skills for interacting with the environment grows, so does the richness and diversity of perceptual experiences. Thus, the learning problem at any given time is constrained by perceptual and motor limitations: learning is more constrained in the beginning of life, and it gradually becomes more flexible across development.

Motor development shapes the perceptual experience of a language learner by determining the set of exploratory actions (e.g. grasping big simple-shaped objects or interacting with small items) and perceptual perspectives (e.g. sitting or standing position) available to the child at any given moment. For example, in the earliest stages of postnatal development, children can only look at objects and people from a distance, which makes big, salient, and frequent items the main data available for learning. Later, children learn to grasp objects and move, which enables them to approach attractive items and manipulate them, generating variable samples that predict their word learning success (Slone, Smith, & Yu, 2019). A large dataset of children's visual views acquired through head cameras shows that the object distributions in children's worldview are right-skewed, meaning that there is a very small number of objects that are almost constantly present in the child's environment (Clerkin et al., 2017); as a result, the names of these objects tend to be learned first and can help bootstrap learning of other words (Kachergis, Yu, & Shiffrin, 2017; Smith et al., 2018). Motor development causes environmental adaptations to child's growth. The most frequent objects in the natural right-skewed distributions of children's visual world change through development: from faces to hands, from rattles to footballs, from small food items available for early grasping to bigger ones. Several potential benefits of such learning distribution are hypothesized (i.e. it might allow consistent learning of gradually changing high-frequency items or support disambiguation of the names of low-frequency items, because they are likely to be surrounded by the already acquired high-frequency items; for more discussion, see Smith et al., 2018; Jayaraman, Fausey, & Smith, 2015; Yoshida & Smith, 2008; Fausey, Jayaraman, & Smith, 2016), but one aspect is undeniable: early word learning is oriented towards a very small number of representative objects which gradually change over time and potentially bootstrap learning of the other objects and their names.

The way that infants sample their learning data is dynamically constrained by their sensorimotor development and their environment, upon which the exploration processes operate. In contrast, modern communicative AI systems are not endowed with developing bodies that constrain their learning within adaptive environments (Fig.3). Rather, in a typical machine language learning task, the learning problem remains constant during training and often represents a desirable end-point, such as answering natural language questions. Moreover, while the communicative machines might learn through weight changes (i.e., in neural networks), their morphology, neural architecture and action space typically remain static across the training. This is analogous to expecting a prelinguistic human baby to learn typing answers to abstract natural language questions from scratch, and then being surprised that the baby takes too much time and too many training examples to learn accurately. In contrast, by "growing" the machine's architecture and morphology in adaptive training environments, we can allow the step-wise and scaffolded accumulation of knowledge (e.g. the neural structure "growing" approach has recently been successfully applied for tackling catastrophic forgetting in continual learning: see Li et al., 2019).

|  | before ~6 months | ~6 months — ~1 year | after ~1 year |
|---|---|---|---|
| **Common language input** | 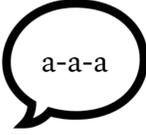 a-a-a | 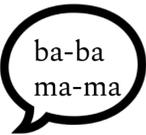 ba-ba ma-ma | 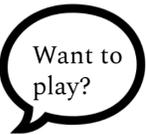 Want to play? |
| **Common objects to interact with** | 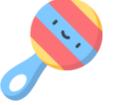 | 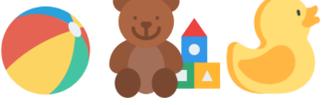 | 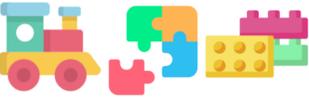 |
| **Motor development** | 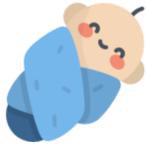 | 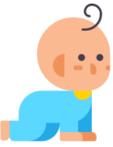 | 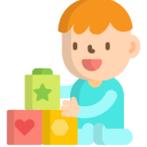 |
| **Action space** | • look around<br>• articulate vowels<br>• cry | • crawl<br>• manipulate big objects<br>• pronounce syllables and simple words<br>• cry | • walk<br>• manipulate big and small objects<br>• combine words in sentences<br>• cry |
| **Common language input** | 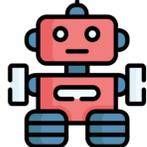 | 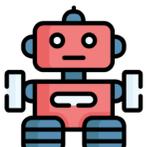 | 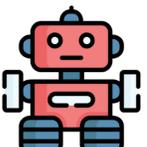 |
| | 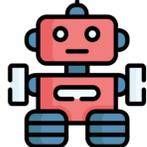 | 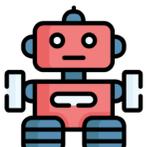 | 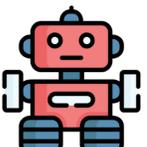 |
| **Action space** | • predict future inputs character by character | • predict future inputs character by character | • predict future inputs character by character |

Figure 3. Interaction of learning environments and individual development. Human babies learn through adaptively constrained interactions with their immediate environments (e.g. objects and agents) which, in turn, adapt to babies' development (top). Machines, in contrast, typically learn in the static contexts that represent the desired end-state of their learning (bottom). Note that the ages, common action space, objects, and language input are approximate.

The figure uses materials from the following sources:

Flaticon.com

Wikipedia contributors. (2021, February 12). Artificial intelligence. In *Wikipedia, The Free Encyclopedia*. Retrieved 21:17, February 14, 2021, from https://en.wikipedia.org/w/index.php?title=Artificial_intelligence&oldid=1006429124

Wikipedia contributors. (2021, February 9). Human intelligence. In *Wikipedia, The Free Encyclopedia*. Retrieved 21:35, February 14, 2021, from https://en.wikipedia.org/w/index.php?title=Human_intelligence&oldid=1005838764

## 3.4. WHERE DO LANGUAGE LEARNING CONSTRAINTS COME FROM?

Domain-general and language-specific constraints guide communicative inferences in infinite spaces of potential solutions, making language acquisition a tractable problem for humans. Natural communicative abilities grow on top of the innate limitations and acquired biases of human perceptual and motor systems (Elman et al., 1996; Regier, 2003; Di Paolo, Cuffari, & De Jaegher, 2018).

Some language learning constraints result from evolutionary processes which adapt species to their evolutionary niche, while other language learning constraints result from developmental processes as an individual gradually learns about the world. Human learner's innate and acquired biases, in turn, determine what forms their communicative systems might take. In the next subsections, I will briefly introduce the ways in which evolutionary, culturally, and individually acquired constraints facilitate human language learning (Fig.4).

Here, again, I focus on the mechanisms that give rise to language learning constraints, rather than on the specific contents of these constraints, which are much harder to determine. This approach contrasts with the "core knowledge" and "core ingredients" approaches in cognitive psychology and AI, which instead focus on the hypothesized high-level contents of the developmental constraints.

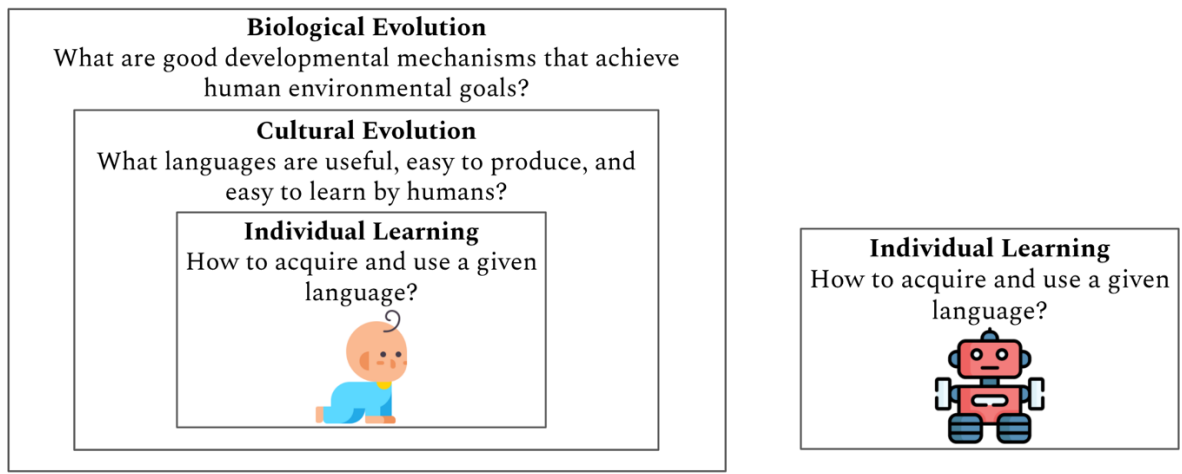

Figure 4. Processes that give rise to human (left) and machine (right) language learning constraints.

The figure uses materials from flaticon.com

### 3.4.1. MACRO-SCALE OF EVOLUTIONARY SEARCH

The biological organization of our species is a precursor to our linguistic abilities and the shape of our languages themselves. Both humans and our communicative systems are not "neutral": human learners inherit a bag of cognitive biases and constraints, and natural languages are well-adapted to them.

Evolution has been adapting our bodies, sensory systems, and learning mechanisms to the ecological niches of our species for millions of years. Most of these adaptations have nothing to do with communication and culture, but these more recent developments are built on top of older components, and are often constrained by them (Elman et al., 1996; Di Paolo, Cuffari, & De Jaegher, 2018). One or a few changes to the pre-existing set of adaptations could have allowed rich and flexible communicative systems to develop in humans (Oudeyer (2006) examines a proof-of-concept computational demonstration, which is also described in section 3.2.1.; Di Paolo, Cuffari, & De Jaegher (2018) provide an overview of the broad existing evidence).

Evolution of language abilities through these curiosity-driven exploration mechanisms, attentional preferences, spontaneous vocal activity, and several other "key" adaptations requires a significantly lower level of evolutionary complexity than the evolution of speech code or language acquisition device (see also Oudeyer & Smith, 2016).

Evolutionary constraints supporting language learning are still under-determined. We know a lot about brain and body organization. However, it is still difficult to trace these low-level components to the development of communication in particular. These biological constraints, however, can often be simulated in artificial neural networks to probe their potential effects on communication learning, which is generally emphasized by biologically-motivated deep learning research programs (e.g. Elman et al., 1996; Marblestone, Wayne, & Kording, 2016; Guerguiev, Lillicrap, & Richards, 2017; Hassabis et al., 2017).

More generally, human cognitive development is optimized on two levels: evolutionary search, on a large scale, explores the space of possible developmental programs for building brains and bodies, while ontogenetic learning, on the individual scale, adapts particular individuals to particular environments using evolutionarily provided mechanisms. Simulations suggest that this combination of searching the space of available learning algorithms (meta-optimization) and individual learning can be very beneficial for finding good reinforcement learning algorithms for a given environment; in many cases, the discovered learning strategies are more successful or more diverse than the algorithms hand-crafted by human engineers (Pourchot & Sigaud, 2018; Houthooft et al., 2018; Khadka & Tumer, 2018; Leite, Candadai, & Izquierdo, 2020). On the other hand, large-scale genetic search facilitates individual learning by finding a good *starting point* to bias the learning processes in directions that are more promising for given tasks or environmental challenges (e.g. Nolfi, Parisi, & Elman, 1994; Todd, Candadai, & Izquierdo, 2020). Thus, AI scientists do not need to wait until all human biases that support communication learning are determined and translated to the language of artificial neural networks to take advantage of the power of optimization performed at the level of learning programs or architectures. For such simulations to be informative, special emphasis should be put into the construction of naturalistic environments where communication can be useful.

While the evolutionary constraints contributing to human language learning are being determined, AI research can benefit from searching for such constraints. General principles, emphasized in the previous sections, as well as considerations of the biologically and environmentally plausible pressures for developing communication, should be combined in the search for innate biases which can successfully guide communication learning in machines.

### 3.4.2. MESO-SCALE OF CULTURAL EVOLUTION: FITTING LANGUAGES TO LEARNING AND SENSORIMOTOR BIASES

What we learn and how we learn it is deeply grounded in our cultural advances. While some of our biological biases facilitate language learning in naturalistic conditions, our languages are fit to our biological constraints through cultural evolution (Christiansen & Chater, 2008; Gibson et al., 2019). Languages in their modern state, artifacts and events that languages refer to, language education, methods for helping children with special needs to acquire language, and many others are the products of human cultural adaptation.

The cultural evolution of language operates on a relatively short timescale of communicative interactions within and between generations. Words that are easy to forget, hard to acquire, or difficult to pronounce are likely to be replaced with other forms (e.g. Nowak & Krakauer, 1999; Nowak, Komarova, & Niyogi, 2002). Not only particular words and word combinations, but also whole language properties adapt to the cognitive limitations of their users. For example, the analysis of morphological complexity of more than 2000 languages showed that languages spoken by larger groups of people tend to have simpler inflectional morphology than languages spoken by smaller groups. This pattern suggests that widely spoken languages adapt to the cognitive limitations of adult second-language learners, who constitute a considerable portion of their speakers (Lupyan & Dale, 2010). Thus, the cognitively constrained communicative interactions ultimately result in communicative systems that are biased to fit their users' predispositions and constraints.

Some language properties arise from the properties of naturalistic communication and learning contexts. For example, communicative systems that are transmitted iteratively – from one generation to another – with a minimal individual experience bottleneck result in the development of compositional concepts (e.g. Kirby, 2001; Kirby & Hurford, 2002; Kirby, 2002; Kirby, Cornish, & Smith, 2008). In particular, when "teachers" cannot supervise learners on every object name or if there is a limit on how many words learners can memorize, then the concepts that refer to a diagnostic property (e.g. color, size or shape) of many objects are maintained, while the identifiers of unique items (e.g. a word that refers to a big green triangle) die out. Thus, the properties of communicative systems are adapted both to the biological constraints of their users and to the properties of naturalistic communicative interactions.

Human languages are optimized for human learning, perception, and production, instead of being universally convenient and learnable. Artificial systems not characterized by these same limitations will not benefit from, and might even be misguided by, the culturally adapted natural languages.

### 3.4.3. MICRO-SCALE OF INDIVIDUAL LEARNING

Individual domain-general and language-specific experiences constrain language learning throughout individual lifetime, and these constraints appear to contribute to human language learning efficiency (e.g. Elman et al., 1996; Seidenberg & MacDonald, 1999; Regier, 2003). Computational simulations of statistical learning of words demonstrate that the gradual accumulation of learning constraints leads to speed shifts in language acquisition (language spurts) (Elman et al., 1996; Regier et al., 2001).

Some constraints emerge from statistical learning of linguistic patterns, while others emerge from perceptual adaptations. For example, simple associative learning quickly gives rise to the mutual exclusivity constraint that greatly accelerates word acquisition (Regier et al., 2001). On the other hand, human attention and perceptual encoding processes flexibly adapt to the cognitive goals of individuals and the statistics of their environments, allowing humans to primarily focus on informative and relevant features at the moment (this perspective is summarized in Goldstone, Leeuw, & Landy, 2015). For example, an individual's auditory perception quickly tunes to the phonetic structure of their language, accentuating the perceptual differences between different phonemes and making the perceptual differences within one phoneme less noticeable (Kuhl et al., 1992; Kuhl, 2000). Thus, human perceptual systems quickly tune to the most cognitively-relevant information in the environment, greatly reducing the amount of less relevant information in perceptual data.

Experience-based constraint accumulation is the most natural domain for connectionist architectures. Designing ecologically valid datasets, training modes, and learning environments (i.e. providing multimodal naturalistic communication data as described in section 3.1; allowing the agents to actively explore their environments as described in section 3.2, and placing the agents in adaptive and controllable communicative environments as described in section 3.3) will help to make sure that the individually learned constraints translate to desired situations of applying this knowledge. Moreover, architectures with adaptive perception (e.g. having active vision or covert attention) are the best candidates to learn perceptual constraints that could accelerate their language learning.

## 4. BUILDING GROUNDED COMMUNICATIVE INTELLIGENCE: REVIEW OF THE DIRECTIONS

### 4.1. GROUNDED LANGUAGE LEARNING: OVERVIEW OF THE PRINCIPLES

As we have seen, there are many ideas from the mind sciences on how "grounding" in adaptive, social, and multimodal environments allows humans to efficiently acquire languages. I suggest that the AI community can potentially benefit from these insights. The language "grounding" ideas that I covered in this paper overlap, complement, and sometimes contradict each other. Therefore, I provide a list of directions for AI researchers, each of which can either be pursued separately or in combination with other directions, depending on the goals of the researcher:

1. **Agents interact with their environments**. Embodied agents interacting with their environments (real or virtual) are more promising candidates for learning human-like language abilities than passive pattern-matching systems. Embodiment allows agents to discover various ways to achieve their communicative goals through sensorimotor shortcuts: by actively manipulating their own perceptual inputs through motor activity.

2. **"Naturalistic" environments**. By using more ecological environments, a developer can ensure that the information that humans use for efficient communication is also available to machines. Careful experimentation with different environments can help to isolate particular experiences or data properties that are crucial for language learning and communication.

3. **Open-ended controllable environments**. The structure of the environment should provide opportunities for continuous interaction. Controllable environments allow the learners to continuously reorganize their learning conditions in the ways that facilitate their language learning process and help them solve communicative tasks. Such possibility enables agents to achieve more distant cognitive goals than the ones which are attainable in the "initial" states of environment provided by the engineers.

4. **Powerful information-seeking exploration mechanisms**. Agents operating in rich, open-ended environments require powerful domain-general exploration mechanisms to guide their learning. Curiosity-driven exploration endows autonomous agents with individualized exploration strategies: such learning can gradually proceed from simpler skills, objects, and concepts to more difficult.

5. **Motor and perceptual development**. Agents who change their motor and perceptual organization across time learn in the hierarchical way: the agent's learning objective, available actions, and data can dramatically transform after they reach a particular developmental milestone. Developing agents can be challenged only with subsets or simplified versions of the full language learning problem at each moment, instead of being required to develop adult-level language capacities from the very start of their learning.

6. **Evolved or naturalistic bodies and neural architectures**. Trajectories of motor and perceptual development that support learning goals in a given environment can be found through carefully designed evolutionary simulations. Evolving the bodies of artificial agents together with their brains (neural architectures) in the target environment enriches optimization possibilities, opening up new ways to find the learning biases and developmental programs that help agents achieve their communicative goals. The analogues of human bodies and their development can also be explored for machines to help them operate in naturalistic environments and succeed in "human-like" communication tasks.

7. **Social learning.** Data and feedback provided by the communication "experts" (e.g. human adults) can guide learner's development in more "objectively" promising directions.

8. **Learning in adaptive social contexts.** Learning in a context of adaptive communicative partners allows agents to continuously receive individualized communicative feedback. Moreover, experience with naturalistic communicative feedback prepares a learner to participate in naturalistic communicative interactions with humans, which are constantly shaped by partners' adaptations to each other.

9. **Multi-agent communicative settings**. Interactive, multi-agent learning environments should be a (distant) goal in building artificial communicative intelligence. Such settings open possibilities for automatized adaptive language teaching and the cultural development of communicative systems that fit particular agents' goals and limitations.

10. **Perceptual simulations**. Embodied models with predictive (or other model-based) feedback can be explored as candidates for learning "perceptual simulations" for language comprehension. The multimodal, rather than purely linguistic, semantic representations are essential for building machines that understand languages in the same way as humans do it.

## 4.2. LIMITATIONS

There are considerable difficulties associated with adopting the grounded approach to language learning in AI. First of all, grounded language learning requires developing new tools, models, environments, and benchmarks, whereas the resources for efficiently training language models in amodal contexts are plentiful. Secondly, simulating the agents that learn to communicate by interacting with their (virtual or real-world) environments might require more time and computing resources than amodal supervised training: the data have to be richer, the adaptive environments and communicative partners have to be simulated or accessed in real time. Thirdly, as the modern state-of-the-art language models might be unadapted for actively exploring and picking up the regularities in interactive multimodal contexts, the agents learning languages in the grounded way might need to undergo substantial development to reach the impressive performance of their amodal counterparts. I hope that these challenges, however, will not dissuade the AI researchers from exploring and gradually incorporating the grounded learning principles into the development of human-like communicative AI.

## 4.3. A CASE OF SUCCESSFUL APPLICATION

Recent work from Hill and colleagues at DeepMind illustrates the potential power of using the language grounding principles for building artificial language learning systems. The researchers trained agents embodied in virtual environments to follow verbal instructions, such as "put a pencil on the bed" or "pick up the dax". The first set of experiments (Hill et al., 2019) demonstrates that active, egocentric perception plays a central role in agent's success in compositional generalization: the ability to understand new combinations of already learned concepts. When agents were learning from their perceptuomotor interactions with the environment through a bounded visual frame (field of view), they developed almost perfect generalization abilities, contrastively to the agents without active bounded perception. The next study showed that when the embodied agents were tested on their ability to follow instructions with entirely new objects, they were able to learn from a single demonstration, showing a capacity known as "fast-mapping" (Hill et al., 2020). Here, the curiosity-driven exploration, rich perceptual experiences, and the temporal properties of the training data played a critical role in allowing the agents to succeed in this challenging fast-mapping task. Thus, preliminary grounded language learning experiments provide existence proofs that embodied agents can solve the same kind of language learning challenges that young human children begin solving early in development. I hope that these new results, combined with evidence from 4E research in cognitive science, will inspire more work in grounded language learning in AI.

## Acknowledgements

I would like to thank Justin Wood and Robert Goldstone for their extremely helpful feedback at every stage of the preparation of this manuscript. I am also grateful to Arseny Moskvichev, Denizhan Pak, Thomas Gorman, Samantha Wood, Ann-Sophie Barwich, Vili Hätönen, Roman Tikhonov, Madeleine Ransom, Peter Todd, Abe Leite, and the participants of the PCL, EASy, and Building a Mind lab meetings for their valuable comments on the earlier versions of this manuscript.